\documentclass{article} 
\usepackage{iclr2027_conference,times}

\usepackage{amsmath,amsfonts,bm}

\def\eqref#1{equation~\ref{#1}}

\def\1{\bm{1}}

\DeclareMathAlphabet{\mathsfit}{\encodingdefault}{\sfdefault}{m}{sl}
\SetMathAlphabet{\mathsfit}{bold}{\encodingdefault}{\sfdefault}{bx}{n}

\usepackage{xcolor}
\usepackage{hyperref}
\usepackage{url}
\usepackage{cleveref}
\usepackage{graphicx}
\usepackage{booktabs}
\usepackage{multirow}
\usepackage{makecell}
\usepackage{enumitem}
\usepackage{wrapfig}
\usepackage{caption}
\usepackage{xspace}

\newcommand{\eg}{\emph{e.g.},\xspace}
\newcommand{\ie}{\emph{i.e.},\xspace}

\definecolor{myDarkBlue}{RGB}{0, 51, 153}
\definecolor{myDarkGreen}{RGB}{0, 102, 51}
\definecolor{myDarkRed}{RGB}{153, 0, 0}
\hypersetup{
    colorlinks=true,
    urlcolor=myDarkBlue,    
    citecolor=myDarkGreen,  
    linkcolor=myDarkRed,    
}

\title{Code Plans, Diffusion Renders: Open-Ended Generative World Modeling}

\author{\\
\textbf{Zixun Fang$^{1,2}$\quad Yawen Shao$^{1,2}$\quad Kai Zhu$^2$\quad Jie Xiao$^2$\quad Shihan Chen$^1$\quad Yu Liu$^2$} \\ \textbf{Xueyang Fu$^1$\quad Yang Cao$^1$\quad Wei Zhai$^1$\quad Zheng-Jun Zha$^1$} \\
\\
$^1$USTC\quad $^2$TongYi Lab\\
}

\crefname{figure}{Fig.}{Figs.}
\crefname{table}{Tab.}{Tabs.}
\crefname{section}{Sec.}{Secs.}

\iclrfinalcopy 
\begin{document}
\lhead{}
\renewcommand{\headrulewidth}{0pt}

\maketitle
\vspace{-0.2cm}
\begin{figure}[!h]
    \centering
    \includegraphics[width=0.8\linewidth]{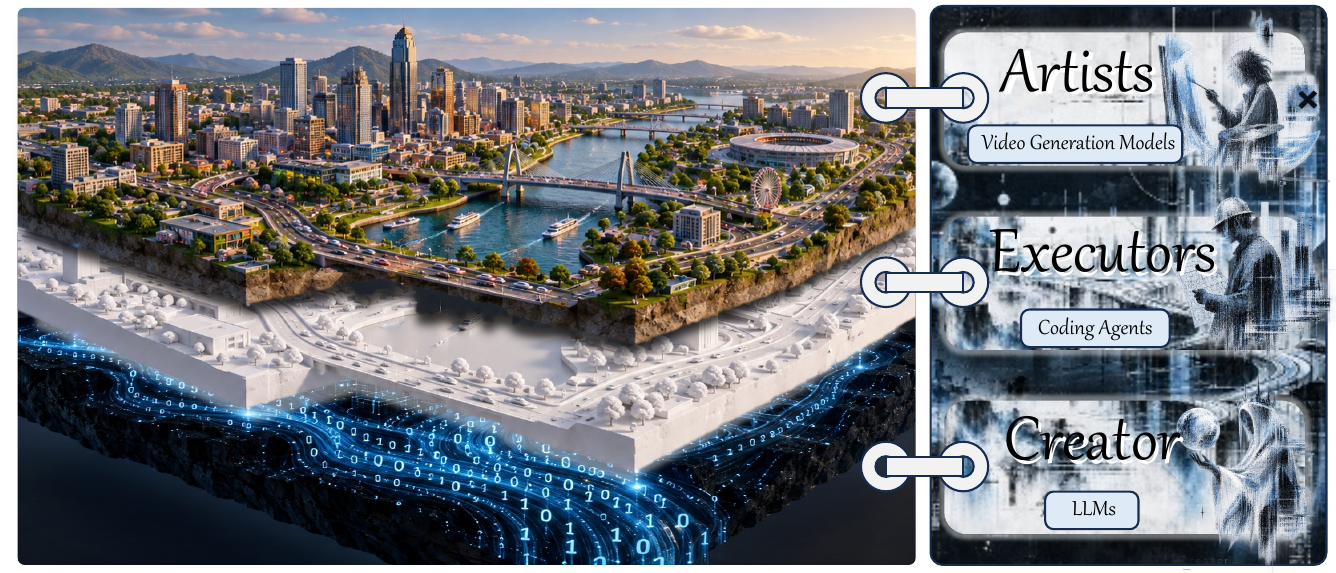}
    \vspace{-0.2cm}
    \captionsetup{font=small}
    \caption{Our conception of a world: the Creator establishes its rules, the Executors build the world accordingly, and the Artist brings it to life.}
    \label{fig:teaser}
\end{figure}

\begin{abstract}
We introduce \textbf{CoDeR}, a new paradigm for world modeling. Unlike existing video world models that implicitly represent world dynamics through visual observations, our system explicitly constructs an executable world with code and employs video generation models for visual realization. Specifically, we coordinate five complementary roles to translate high-level concepts into structured world rules, executable dynamics, and perceptual observations. This design enables \textit{long-term memory}, \textit{open-ended interactions}, \textit{autonomous world evolution}, and \textit{multi-agent scenarios}, where multiple entities can act, interact, and evolve persistently beyond the current observation. Extensive experiments demonstrate that our framework substantially extends the capabilities of existing world models, enabling long-term memory, open-ended interactions, autonomous evolution, and persistent multi-agent dynamics, while achieving state-of-the-art performance across multiple evaluation settings. Code and model weights will be made publicly available.
Project Page: \href{https://becauseimbatman0.github.io/CoDeR}{CoDeR}.

\end{abstract}

\section{Introduction}

World modeling aims to simulate interactive environments that evolve in response to agents’ actions, with applications in embodied AI~\citep{wm_cosmos3_2026,wm_nwm_2024}, video games~\citep{wm_mineworld_2025,wm_realplay_2025,wm_diamond_2024}, and virtual reality~\citep{wm_unisim_2023,wm_generated_reality_2026}. Many recent approaches build on video generation models to synthesize visually compelling observations~\citep{wm_viewcrafter_2024,wm_cinescene_2026,wm_neoverse_2026,wm_moverse_2026}, incorporating control signals~\citep{wm_gamefactory_2025,wm_hunyuan_gamecraft_2025,wm_yume_2025,wm_the_matrix_2024} such as camera poses to enable interaction~\citep{wm_cameractrl_2024}. However, these approaches often encode world state implicitly in visual histories, which can make it difficult to maintain persistent memory, model off-screen dynamics, and enforce consistent object interactions~\citep{ma2026outofsight,wm_worldmodelbench_2025}. Moreover, visual observations reveal only part of a world: underlying states, rules, and relationships—such as resource ownership, physical constraints, and social connections~\citep{wm_generative_agents_2023}—can shape future events without being directly visible.

A coherent world should evolve according to well-defined rules, even when it is not being observed. This principle reflects everyday experience: water placed in a functioning freezer continues to cool and eventually freezes. Its evolution depends on physical conditions, rather than its visibility to an observer. This motivates a world representation that maintains persistent state~\citep{wm_persist_2026,wm_lsm_world_2026} and governs its evolution independently of visual observation. Meanwhile, recent large language models (LLMs) exhibit remarkable world knowledge and robust reasoning capabilities, offering a promising route toward this goal: translating a world into executable rules and state transitions~\citep{wm_worldcoder_2024,piriyakulkij2025poeworld}. Such a formulation enables the maintenance of the underlying world representation, while a video generation model synthesizes temporary visual observations on top of the structured world~\citep{wm_magpie_2026}.

Realizing this vision requires more than generating executable scene code. A complete world comprises heterogeneous systems, such as transportation, construction, and resource management, each governed by local mechanisms while remaining subject to shared constraints. Constructing these systems therefore requires a way to decompose the world into manageable components and coordinate their interactions. Equally important is the connection between executable state and visual observation: abstract rules and state transitions must be translated into spatial and temporal conditions that a video generation model can follow~\citep{wm_dar_2026}. These challenges call for a framework that organizes world construction and connects its execution to visual synthesis.

To this end, we introduce CoDeR, a framework organized around five complementary roles: the God of Concepts, the Creator, Executors, Artists, and Travelers. Specifically, the God of Concepts expresses the desired world through language or images. The Creator, instantiated as a large language model, interprets this intent, elaborates shared rules and a thematic direction, and produces a world design blueprint. Executors, a group of coding agents, translate the blueprint into executable entities, behaviors, and systems. Artists, implemented as video generation models, transform visual conditions derived from this executable world into detailed artistic realizations. Finally, Travelers explore the environment and experience its unfolding events from their own perspectives. Together with our proposed Logical Spaces strategy and Observation as World Registration paradigm, CoDeR achieves consistent world generation, open-ended interactions, dynamic evolution and multi-agent collaboration.

In summary, our contributions are as follows:
\begin{itemize}[leftmargin=1.2em]
\item We introduce \textbf{CoDeR}, a new paradigm
for world modeling that coordinates five complementary roles
to transform conceptual intent into executable, evolving,
and visually expressive worlds.
\item We propose \textit{Logical Spaces} to organize collaborative
world construction under shared rules, and
\textit{Observation as World Registration} to integrate generated
observations into a persistent world representation, enabling
visual information to be retained and reused across viewpoints
and interactions.
\item Extensive experiments demonstrate the superior performance
of CoDeR across key world modeling capabilities,
ranging from persistent memory to multi-agent interaction.
\end{itemize}

\section{Related Work}
\subsection{Video Generation} 
Extending image generation to the temporal domain, video generation
models aim to synthesize high-quality, temporally coherent videos~\citep{wm_cosmos_platform_2025}
through training on large-scale video datasets~\citep{wm_cosmos_predict_transfer25_2025}.
For models trained with flow matching~\citep{lipman2023flowmatching}, a neural network learns
a velocity field along a prescribed path between the data and
noise distributions.
Formally, let $x_0$ denote a clean video sample or its latent
representation, $c$ the associated conditioning information,
and $\epsilon \sim \mathcal{N}(0,I)$ Gaussian noise.
A linear interpolation between data and noise is defined as:
\begin{equation}
    x_t = (1-t)x_0 + t\epsilon,
    \qquad t \in [0,1].
    \label{eq:flow_path}
\end{equation}
The model $v_\theta$ is trained to predict the target velocity
$\epsilon-x_0$ by minimizing:
\begin{equation}
    \mathcal{L}_{\mathrm{FM}}
    =
    \mathbb{E}_{x_0,c,\epsilon,t}
    \left[
        \left\|
            v_\theta(x_t,t,c) - (\epsilon-x_0)
        \right\|_2^2
    \right],
    \label{eq:flow_loss}
\end{equation}
where $(x_0,c)$ is sampled from the training distribution and
$t \sim \mathcal{U}(0,1)$.
At inference, samples are generated by numerically integrating
the learned velocity field backward from $t=1$ to $t=0$,
starting from Gaussian noise.

Beyond visual quality, controllable video generation~\citep{wm_hunyuan_gamecraft_2_2025,wm_shadowdancer_2026,wm_incantation_2026} aims to provide
fine-grained control over the content and dynamics of synthesized
videos~\citep{wm_videocomposer_2023}.
Within the above formulation, the conditioning information $c$ can
incorporate structured signals such as human poses~\citep{wm_lome_2026,wm_hand2world_2026,wm_anchorworld_2026,wm_playerone_2025}, camera
trajectories~\citep{wm_aether_2025}, and scene layouts~\citep{wm_dwm_2025,wm_cosmos_transfer1_2025}.
These signals guide the generation process toward desired spatial
configurations and temporal behaviors~\citep{wm_scenescape_2023,wm_stargen_2025}.

\subsection{World Modeling}
Beyond passive video synthesis, world models aim to simulate an environment as a persistent process whose state changes in response to actions, events, and the passage of time. Recent advances in generative modeling~\citep{wm_wonderworld_2024,wm_omnix_2025,wm_lyra_2025} have substantially improved the visual fidelity of such simulated worlds~\citep{wm_voyager_2025,wm_deepverse_2025}, while progressively extending them toward interactive, persistent, and autonomous environments.

\textbf{Interactivity.} A fundamental capability of world models is to respond to actions~\citep{wm_scope_2026,wm_egohoi_2026,wm_impact_2026} rather than merely generate a predetermined visual trajectory. Early works such as Genie~\cite{bruce2024genie} learn action-controllable environments from large-scale videos, while GameNGen~\cite{valevski2024gamengen} demonstrates that a diffusion model can directly serve as a real-time neural game engine. Subsequent approaches, including GameGen-X~\cite{che2024gamegenx}, Matrix-Game~\cite{zhang2025matrixgame}, and its real-time extensions~\citep{wm_matrix_game_2_2025,wm_matrix_game_3_2026}, further improve action controllability, visual quality, and streaming efficiency~\citep{wm_worldplay_2025,wm_wonder_2026,wm_matrix_game_35_2026}. Nevertheless, the action spaces of many existing world models remain dominated by navigation or predefined control signals. Recent methods such as ActWorld~\cite{xiong2026actworld} begin to support richer mid-rollout object interactions, highlighting the transition from merely \emph{explorable} environments toward genuinely \emph{interactive} worlds~\citep{wm_lingbot_world_2_2026,wm_yume_15_2025,wm_dreamx_world_2026}.

\textbf{Memory.} Long-term interaction further requires a world to preserve information beyond the immediate generation context~\citep{wm_memory_forcing_2025,wm_vrag_2025,wm_lyra2_2026,wm_closing_the_loop_2026}. This includes not only temporal continuity, but also persistent object identities, spatial layouts~\citep{wm_spmem_2025,wm_mosaicmem_2026,wm_anchorweave_2026}, and the consequences of previous interactions when a location is revisited. Recent world models therefore increasingly incorporate explicit long-horizon memory mechanisms~\citep{wm_worldmem_2025,wm_spatia_2025,wm_gen3c_2025}. RELIC~\cite{hong2025relic}, for example, compresses historical observations into camera-aware latent memories for real-time long-duration exploration. Related approaches such as AlayaWorld~\cite{alayaworld2026} integrate compressed history and geometry-aware spatial memories to stabilize long autoregressive rollouts. These efforts substantially extend the effective temporal horizon of video world models~\citep{wm_memlearner_2026,wm_worldpack_2025,wm_reworld_2026}; however, such memory is primarily designed to reconstruct or retrieve previously observed states~\citep{wm_vmem_2025,wm_context_as_memory_2025}, rather than to explicitly model how the underlying world itself changes over time.

\textbf{Evolution.} A persistent world should not only remember its past, but also continue to evolve independently of the observer. This distinction exposes a fundamental limitation of observation-centric video world models: when an entity leaves the camera view, its internal state may effectively stop evolving until it becomes visible again. Recent studies explicitly identify this \emph{out-of-sight dynamics} problem. LiveWorld~\cite{duan2026liveworld} addresses this problem by separating observation rendering from a persistent global state, allowing dynamic entities to continue evolving while they are outside the current field of view. ReMind~\cite{xu2026remind} further trains video generators to retrieve and propagate hidden dynamic states across observation gaps. These approaches move world modeling beyond static spatial memory toward persistent temporal processes~\citep{wm_flowm_2026,wm_hydra_2026}. Nevertheless, supporting open-ended evolution---where independent entities, events, and processes can autonomously alter the world over arbitrarily long timescales---remains largely unexplored.

\textbf{Beyond Vision.}
%
More fundamentally, a world is not merely a sequence of visual observations. Pixels describe how a world \emph{appears}, but do not explicitly represent the concepts, rules, relations, and causal mechanisms that determine how it operates. This has motivated recent efforts to augment neural world models with structured and executable representations~\citep{wm_worldcoder_2024,wm_chronoagentic_2026,piriyakulkij2025poeworld}. Agent World Model~\cite{wang2026agentworldmodel} constructs code-driven, database-backed environments for training interactive agents, providing explicit and reliable state transitions beyond natural-language simulation. Collectively, these works suggest a transition from purely visual world models toward hybrid systems in which structured representations govern world logic and generative models realize perceptual observations~\citep{wm_mass_2026,wm_marionette_2026}. Our CoDeR follows this direction while further organizing world construction and evolution through multiple specialized agents, enabling world logic, autonomous processes, and visual realization to evolve collaboratively rather than being represented by a single monolithic visual dynamics model.

\section{CoDeR}
\subsection{Overview}
%
%
%
This section presents the pipeline of our proposed CoDeR. We begin by elaborating on the functionalities of the roles introduced within the system, followed by an illustration of how these roles interact and collaborate to construct an interactive and continuously evolving world model.

\textbf{The God of Concepts.} ``Let there be light!'' said the God. 
%
%
%
In our world system, the God of Concepts serves as the ultimate origin of the world to be created. In practice, this role can freely describe the desired world through text or images, in a manner similar to how prior works directly prompt video generation models~\citep{che2024gamegenx}. However, unlike the prevalent paradigm in current world modeling, where a text encoder or prompt enhancer processes this intent before feeding it directly into a DiT (Diffusion Transformer)~\citep{peebles2023dit}, our system applies only minimal modification (\eg format alignment) to the input—motivated by our belief that a world is inherently difficult to describe using only one or a few sentences—and instead forwards this intent to the next node, the Creator.

%
%
\textbf{The Creator.} To faithfully yet creatively realize this will, the Creator establishes the rules and sets the tone for the world. Powered by advanced LLMs, the Creator recursively decomposes and refines the idea through a set of agents, each following a structured duty and collaboratively shaping different aspects of the world—such as its visual appearance and operational mechanics.
%

As is well known, many aspects of a world are difficult to infer directly from visual information alone. Examples include underlying rules (\eg traffic regulations), the internal states of agents (\eg health or reputation), and precise physical dynamics governing motion (\eg Newton's laws). This is a key reason why previous video-generation-based paradigms often fall short in maintaining long-term world consistency and reasoning about such hidden information. Specifically, since these models are trained to directly synthesize pixel-level appearances from data, they lack any explicit mechanism to represent or track such information—information that cannot be readily obtained from visual cues alone.
In contrast, the Creator addresses this challenge by explicitly defining such hidden information—including rules, agent states, and physical dynamics—as part of the world state from the very outset, thereby ensuring that all subsequent generation strictly adheres to these constraints.
In a word, the Creator is the ``main brain'' of the world.

\textbf{Executors.} Once the Creator has established the governing rules, Executors are assigned to complete different parts of the desired world. Driven by advanced multimodal LLMs, Executors act as coding agents that generate executable code to instantiate the whitebox world, \ie a 3D blockout representation constructed entirely through code, in strict accordance with the Creator's blueprint.

Representing the world through executable code, rather than raw pixels, allows the hidden information to be explicitly encoded and enforced, rather than being implicitly and unreliably inferred from visual appearance.

\textbf{Artists.} The world needs art. 
%
%
%
While representing the world through executable code provides precise control over its underlying rules and geometric structure, the resulting whitebox world inevitably lacks visual richness—appearing plain, untextured, and empty. To address this limitation, Artists are introduced to transform the whitebox world into a visually compelling representation, endowing it with a diverse range of visual styles—from photorealistic scenes to anime-inspired aesthetics—all while preserving the structural and logical fidelity established by the Creator and Executors.
%


Specifically, we tame state-of-the-art video generation models to serve as the Artists within this hierarchically constructed world. An Artist observes a segment of the otherwise drab world and innovatively translates it into a visually appealing, detail-rich video aligned with the Creator's tone. After generation, the Artist returns the video to the Executors, which register its appearance in the corresponding world region—a mechanism we refer to as \textit{Observation as World Registration}, which we elaborate on below.

\begin{figure}[t]
    \centering
    \vspace{-1cm}
    \includegraphics[width=0.8\linewidth]{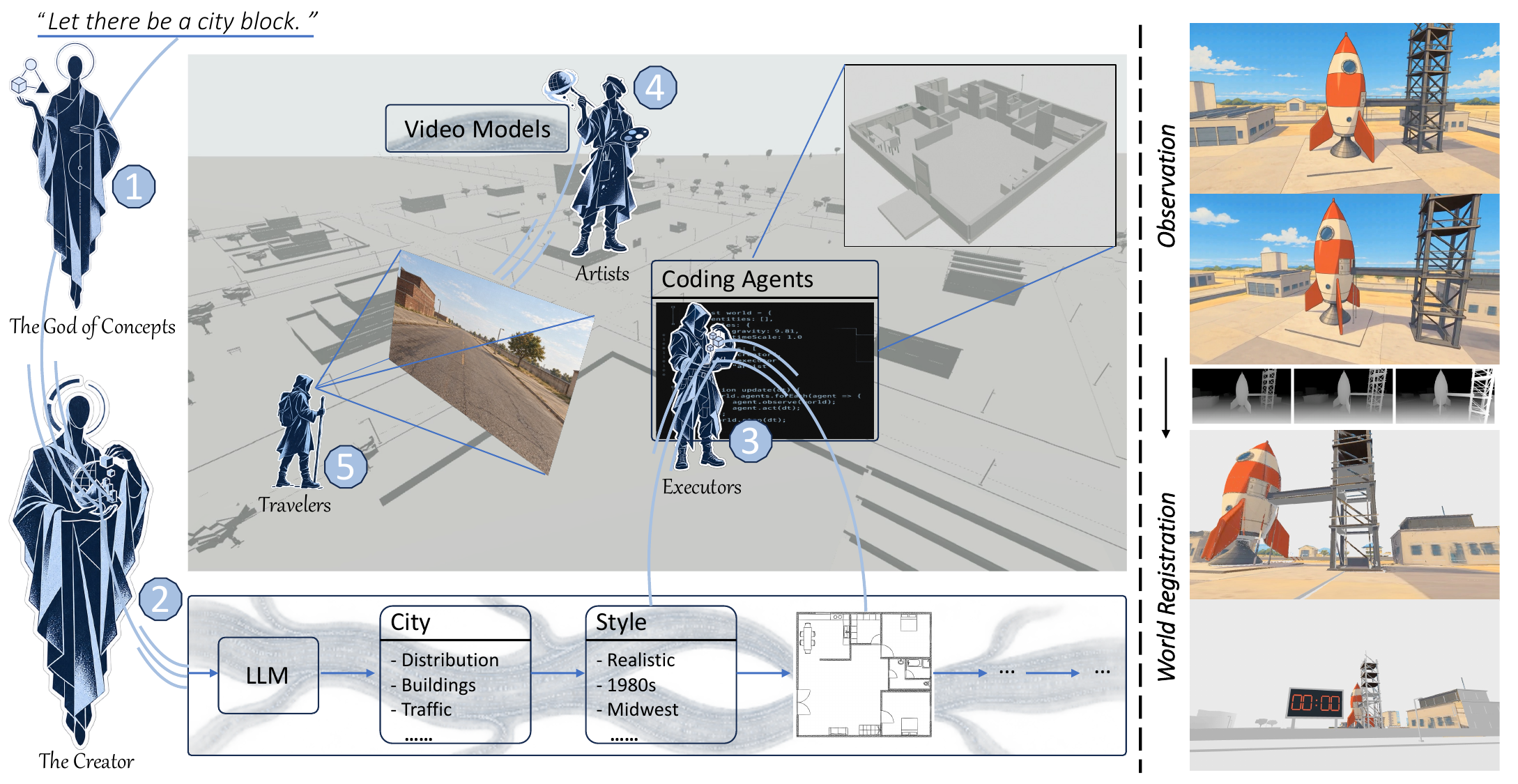}
    \vspace{-0.2cm}
    \captionsetup{font=small}
    \caption{\textbf{Method Overview.} \textbf{Left:} The pipeline of our CoDeR. The God of Concepts conveys its intent to the Creator. The Creator (LLM) orchestrates the entire world, while the Executors (coding agents) implement the whitebox world. Finally, the Artists (video generation models) render observations along the trajectories sampled by the Travelers. \textbf{Right:} The Observation as World Registration paradigm. Once an observation is generated, it is registered back into the world using its depth map.}
    \label{fig:method}
\end{figure}

\textbf{Travelers.} 
%
As the world's explorers, Travelers visit the world, interact with the environment and other entities, and determine which regions require rendering by the Artists as they explore. In this world system, a Traveler may take the form of a visible entity with a defined appearance, or simply exist as a disembodied camera viewpoint.

\subsection{World System}
\textbf{Logical Spaces.}
\label{method: logical spaces}
Once the key parameters of a world have been established by the Creator, our harness decomposes the global construction objective \(g\) into a graph of \textit{\textbf{logical spaces}}:
\begin{equation}
\mathcal{G}=(\mathcal{V},\mathcal{E})=\mathcal{D}(g),
\qquad
\mathcal{V} = \{(g_i,\mathcal{K}_i)\}_{i=1}^{N}
\end{equation}
where \(\mathcal{D}\) denotes the decomposition process, \(\mathcal{V}\) contains \(N\) logical spaces indexed by \(i\), and \(\mathcal{E}\) specifies their dependencies and connections. Each logical space is characterized by a local objective \(g_i\) and an interface contract \(\mathcal{K}_i\), which specifies its inputs, outputs, and construction constraints. These contracts incorporate the shared world rules established by the Creator while leaving space-specific implementation choices to individual Executors. A logical space therefore defines a functional scope rather than necessarily a disjoint spatial region.

Each logical space is assigned to an Executor, \ie a coding agent, which constructs its corresponding module:
\begin{equation}
B_i=\mathcal{A}_i(g_i,\mathcal{K}_i),
\qquad i=1,\ldots,N,
\label{eq:logical_construction}
\end{equation}
where \(\mathcal{A}_i\) denotes the construction process performed by the assigned Executor, including code generation, tool execution, and local refinement, and \(B_i\) denotes the resulting module with its scene elements, executable behaviors, and exposed interfaces. Executors can develop modules concurrently once their interface contracts and required dependencies are available. For example, constructing a traffic system involves vehicle design, traffic regulations, and road network layout, making end-to-end development by a single Executor challenging and time-consuming. Decomposition allows these responsibilities to be distributed across multiple Executors.

The harness subsequently integrates the resulting modules through explicit interface bindings:
\begin{equation}
W = \operatorname{Compose}(
    \{B_i\}_{i=1}^{N},
    \{\operatorname{Bind}_{\beta_{ij}}(B_i,B_j)\}_{(i,j)\in\mathcal{E}}
),
\end{equation}
where \(W\) is the assembled, executable world, \(j\) indexes a connected module, and \(\beta_{ij}\) specifies the binding between modules \(B_i\) and \(B_j\), such as a spatial transformation, state mapping, or event connection. The operator \(\operatorname{Bind}\) instantiates each connection prescribed by \(\mathcal{E}\), while \(\operatorname{Compose}\) assembles the modules and their connections into a unified system. For instance, a vehicle's visual geometry and collision geometry can be developed concurrently under an agreed spatial specification and subsequently bound to the same vehicle state. Similarly, a separately constructed cockpit can expose a driver-camera interface that is bound to the vehicle's pose, allowing its interior viewpoint to observe the shared world. This design supports modular, concurrent construction while preserving explicit relationships among logical spaces.

\textbf{Observation as World Registration.}
\label{method: observation}
To preserve a consistent world state each time an Artist paints a segment of the world, we propose the \textit{Observation as World Registration} paradigm. The core idea is that once an Artist has generated a visual rendering of a segment, this generation is registered back into the world, becoming part of the observation that any subsequent agent perceives when looking at that region.

The Artist model in our system is a video generation model, which takes the drab, plain whitebox observation along with its corresponding depth as input, and outputs a visually rich, colorful video. Given an observation chunk with $n$ frames sampled from the whitebox world, denoted as the video $\mathcal{W}=\{w_0, w_1,..., w_{n-1}\}$, we can readily obtain the corresponding ground-truth depth video $\mathcal{D}=\{d_0, d_1,..., d_{n-1}\}$, as well as semantic information about the sampled location and its surroundings. While not directly discernible from $\mathcal{W}$ itself, it can be directly retrieved from the whitebox world's underlying state, since the identity and attributes of every entity are already known. We then construct a prompt $\mathcal{P}$ based on this information, such that $\mathcal{V}=\text{Artist}(\mathcal{W},\mathcal{D},\mathcal{P})$, where $\mathcal{V}=\{v_0, v_1,..., v_{n-1}\}$ is the resulting generated video. 

To endow the Artist model with the ability to perceive historical context, we introduce a partial registration mechanism. Specifically, given a previously generated video $\mathcal{V}$ and its corresponding ground-truth depth $\mathcal{D}$, we back-project a randomly sampled subset of $\mathcal{V}$'s pixels onto the whitebox world $W$ and render it from the observation viewpoints, yielding a partially registered observation video $\mathcal{R}$—wherein some regions retain their original plain appearance while others have already been colored according to prior generations. 
Concretely, the known camera parameters and world geometry allow us to associate the selected pixels with the corresponding surfaces and project their colors into the observation viewpoints. We retain only projections that correspond to the same surface and pass depth-based visibility checks, blending valid observations where they overlap. Denoting these projected colors by $\Pi(\mathcal{V},\mathcal{D})$, with zeros at uncovered locations, and their binary coverage mask by $\mathcal{M}$, the partial registration is written as:
\begin{equation}
    \mathcal{R}
    = \mathcal{M} \odot \Pi(\mathcal{V},\mathcal{D})
    + (\mathbf{1}-\mathcal{M}) \odot \mathcal{W},
    \label{eq:partial_registration}
\end{equation}
where $\odot$ denotes element-wise multiplication and $\mathcal{M}$ is one in registered regions and zero elsewhere. Thus, previously observed appearance becomes part of the Artist's next observation, while unobserved regions retain the whitebox appearance for subsequent generation. We then train the Artist model to recover the complete, fully colored video from this partial observation.


\section{Experiments}
\subsection{Whitebox World Generation}
\label{exp:whitebox}
We leverage Three.js as the framework for constructing the code-generated whitebox world, owing to its lightweight, programmable, and composable interface.
This allows Executors to construct independent parts of the world as modular code snippets, which can then be seamlessly linked back together to form the integrated world as described above in~\cref{method: logical spaces}.
%
%
For human-related scenarios, we adopt SMPL-H~\citep{romero2017embodiedhands} as the underlying representation to model human motion, hand-object interaction, and viewpoint binding.

We implement Creator and Executors using GPT-6 Astra, an advanced multimodal LLM-based coding agent with strong 3D spatial awareness. Leveraging its capabilities in multimodal reasoning and geometric understanding, GPT-6 Astra is able to interpret spatial constraints, reason about object placement and interaction, and generate executable Three.js code that faithfully reflects the intended design.

\subsection{Artist Model}

We choose MiniMax H3~\citep{minimax2026h3}, a state-of-the-art open-source video generation model, as our Artist model. Although this video generation model can natively re-render whitebox-like videos into colorful ones, we find that it tends to produce render-style outputs—\eg hard edges, monotonous textures, and flat lighting—which undermine the diversity and realism of the generated visuals. In addition, it is difficult to directly adapt this model to our proposed Observation as World Registration paradigm without further tuning.

To address this, we propose a \textit{Visual Cue Hacking} strategy, which extracts common and reliable visual cues (\eg Canny edges and depth maps) from the whitebox world and leverages them to guide the model taming process.

%
%

\textbf{Data Curation.} To construct training data using the proposed Visual Cue Hacking strategy, we begin by sampling a random trajectory within the code-generated whitebox world, and extract the corresponding data following the same procedure described in~\cref{method: observation}. Following the notation introduced earlier, this yields a whitebox video $\mathcal{W}$ together with its corresponding prompt $\mathcal{P}$. We then extract Canny edge maps from $\mathcal{W}$ (depth is directly available as $\mathcal{D}$), and feed these cues into ControlNet~\citep{wm_controlnet_2023} to generate videos spanning diverse visual styles, denoted as $\mathcal{V}$. In this way, we obtain training tuples ($\mathcal{W}$ or $\mathcal{R}$, $\mathcal{D}$, $\mathcal{P}$, $\mathcal{V}$) for our Artist model.

\textbf{Training.} We train our Artist model using 32 NVIDIA A800 GPUs, with a global batch size of 16 for 200 training steps. We adopt AdamW~\citep{loshchilov2019adamw} as the optimizer with a learning rate of $1\times10^{-5}$. Each video clip is resized to a resolution of $1280\times704$ and temporally sampled to contain 124 frames. During training, the transformer backbone is fine-tuned using LoRA~\citep{hu2022lora} while the ControlNet branch undergoes full-parameter fine-tuning.

\begin{figure}[t]
    \centering
    \vspace{-1cm}
    \includegraphics[width=0.8\linewidth]{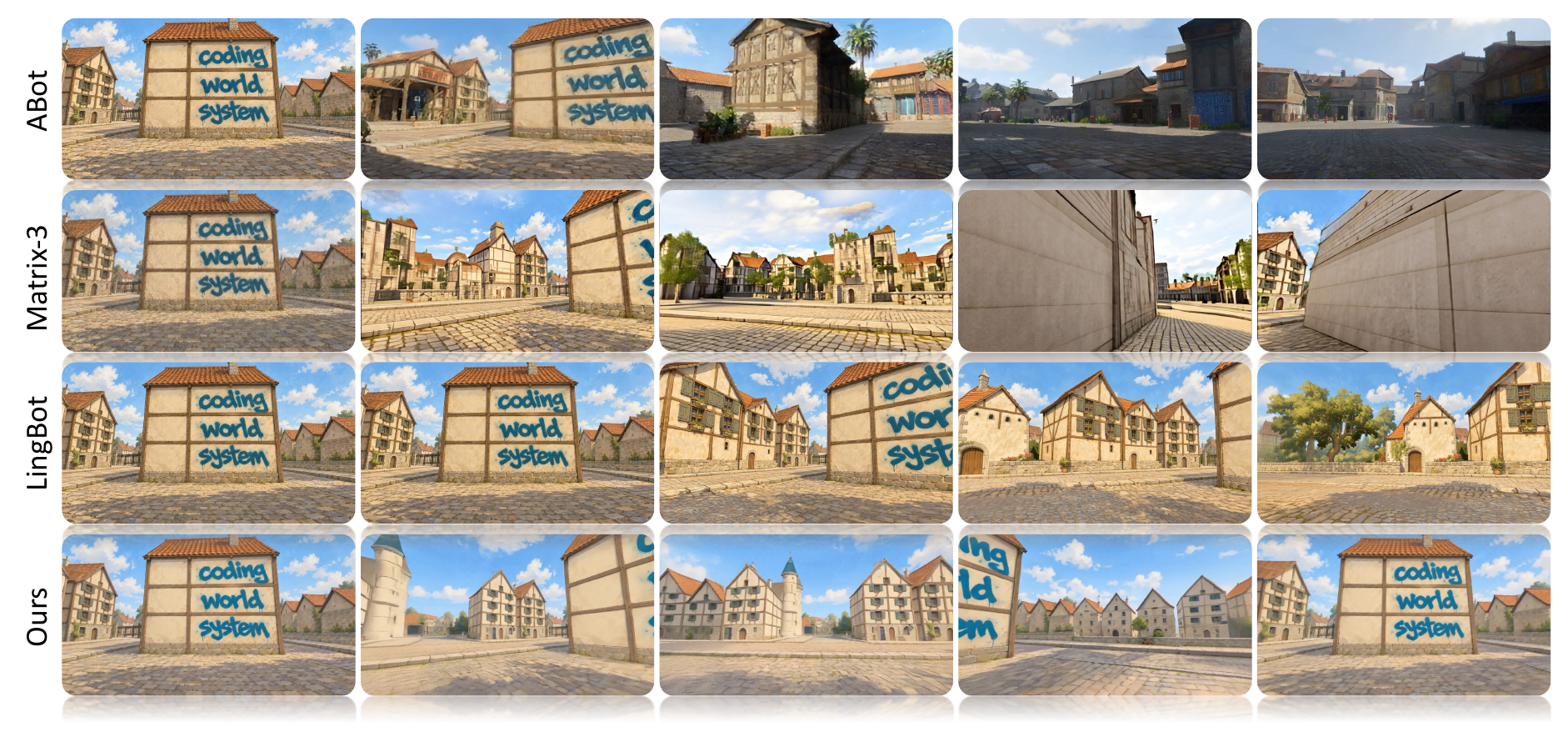}
    \vspace{-0.3cm}
    \captionsetup{font=small}
    \caption{We sample a rotational trajectory to examine whether the ``Coding World System'' mark is consistently maintained on the wall. The results show that ABot-World exhibits severe quality degradation, while Matrix-Game 3.0 fails to preserve the mark. LingBot-World fails to accurately respond to the control signals, resulting in duplicated frames. In contrast, our method successfully follows the rotational trajectory while consistently preserving the mark on the wall.}
    \label{fig:comp1}
\end{figure}

\subsection{Qualitative Comparison}
In this section, we demonstrate that our CoDeR exhibits several key properties that a reliable world should possess, including memory, open-ended interactivity, and continuous evolution, and further explore its capabilities in multi-agent scenarios~\citep{wm_multiworld_2026,wm_prisma_world_2026,wm_metaworld_2026,wm_solaris_2026}.

\textbf{Memory.}
Although our Artist model is trained with a fixed context length of 124 frames, we observe strong long-term memory capabilities enabled by our proposed Observation as World Registration strategy. We evaluate a 360-degree rotation to examine whether the tested methods can preserve the scene structure and the ``Coding World System'' mark over a long temporal horizon. As illustrated in~\cref{fig:comp1}, ABot-World~\citep{wm_abot_world_0_2026} exhibits severe quality degradation, while Matrix-Game 3.0~\citep{wm_matrix_game_3_2026} loses the building structure, and LingBot-World~\citep{wm_lingbot_world_2026} fails to accurately follow the input trajectory. In contrast, our method achieves a closed-loop rollout while preserving both geometric and appearance consistency.

\begin{figure}[t]
    \centering
    \vspace{-1cm}
    \includegraphics[width=0.8\linewidth]{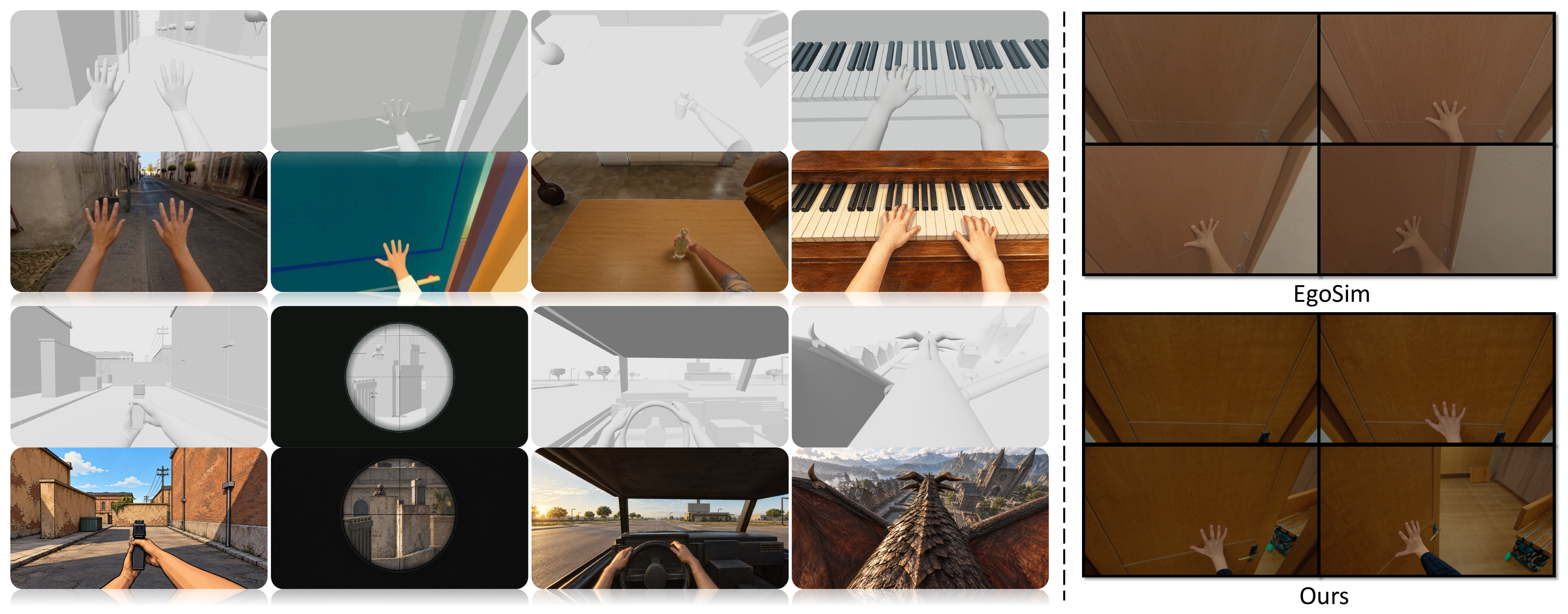}
    \vspace{-0.3cm}
    \captionsetup{font=small}
    \caption{\textbf{Left:} We support open-ended interactions ranging from playing the piano to riding a dragon. \textbf{Right:} Comparison with EgoSim, which fails to open the door, while our method successfully opens it and reveals the new environment.}
    \label{fig:comp2}
\end{figure}

\textbf{Open-ended Interactions.}
We enable open-ended world interactions, ranging from opening a door to piloting a spaceship, by leveraging powerful code-defined interaction logic. Rather than relying on a predefined action space, our CoDeR allows the Creator to define new actions, which are subsequently implemented by the Executors, making the action space continuously extensible. As shown in~\cref{fig:comp2}, the left part of the figure demonstrates diverse code-defined actions in the whitebox world together with their corresponding visual realizations by the Artists. Experiments further show that conventional methods with predefined interactions tend to fail on challenging environment-level interactions, such as opening doors. As illustrated on the right side of~\cref{fig:comp2}, EgoSim~\citep{wm_egosim_2026} is unable to open the door. In contrast, our method successfully opens the door and reveals the new environment behind it.

\begin{figure}[!b]
    \centering
    \vspace{-0.6cm}
    \includegraphics[width=0.8\linewidth]{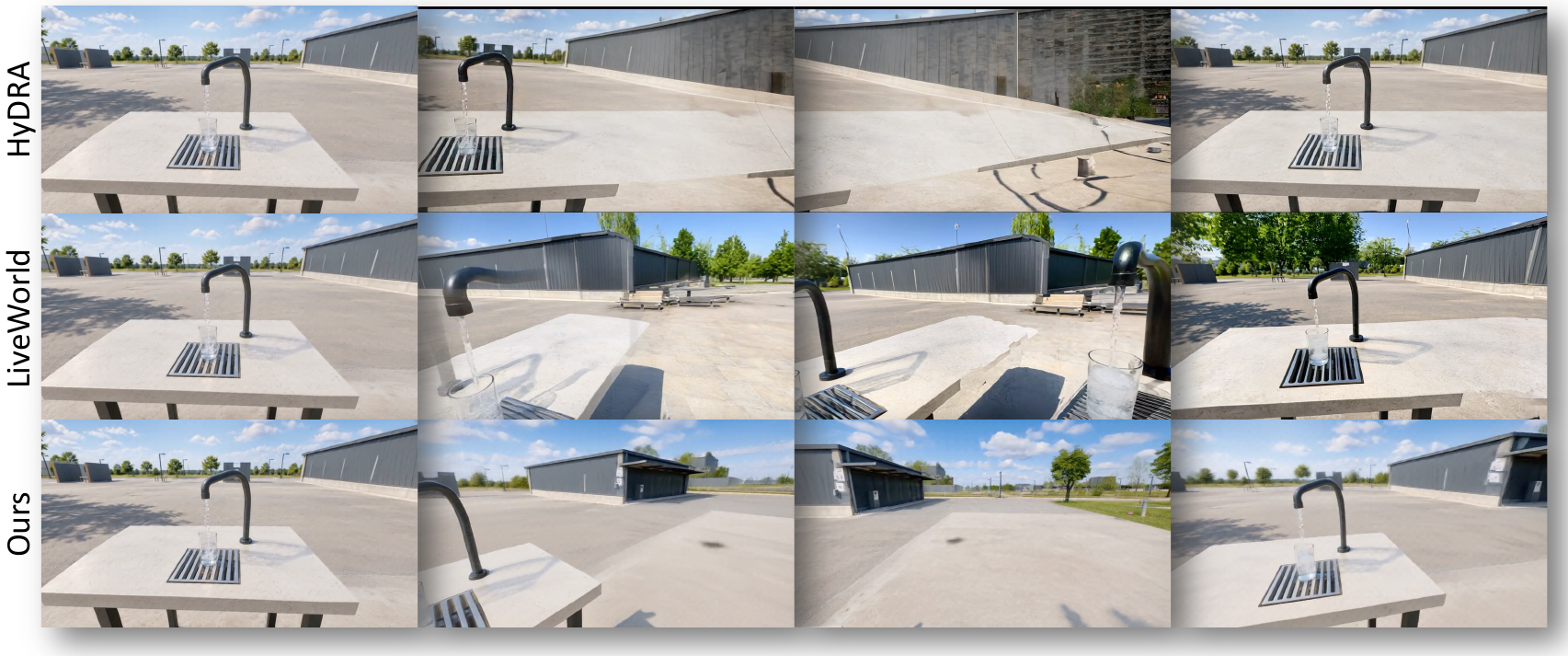}
    \vspace{-0.3cm}
    \captionsetup{font=small}
    \caption{We demonstrate the world evolution mechanism of our method in comparison with other approaches. In the first frame, a tap is pouring water into a glass cup. The camera then moves away from the cup and later returns to examine whether the water level has continued to rise. HyDRA produces a static water column, while LiveWorld exhibits obvious visual artifacts. In contrast, our method continuously updates the water level even when the cup is out of sight.}
    \label{fig:comp4}
\end{figure}

\textbf{Evolution.}
Our CoDeR effectively models event evolution, an essential capability for maintaining persistent dynamics in world models. We evaluate this capability against existing methods in~\cref{fig:comp4}. The results show that HyDRA~\citep{wm_hydra_2026} fails to model the continuous water-pouring process, as the water column remains nearly static across frames. LiveWorld~\citep{duan2026liveworld} captures some water dynamics but exhibits obvious visual artifacts and fails to correctly update the water level in the cup after it moves out of sight. In contrast, our method accurately captures the water dynamics and continuously updates the underlying state even when the cup is outside the field of view, correctly reflecting the increased water level when it becomes visible again.

\begin{figure}[t]
    \centering
    \vspace{-1cm}
    \includegraphics[width=0.8\linewidth]{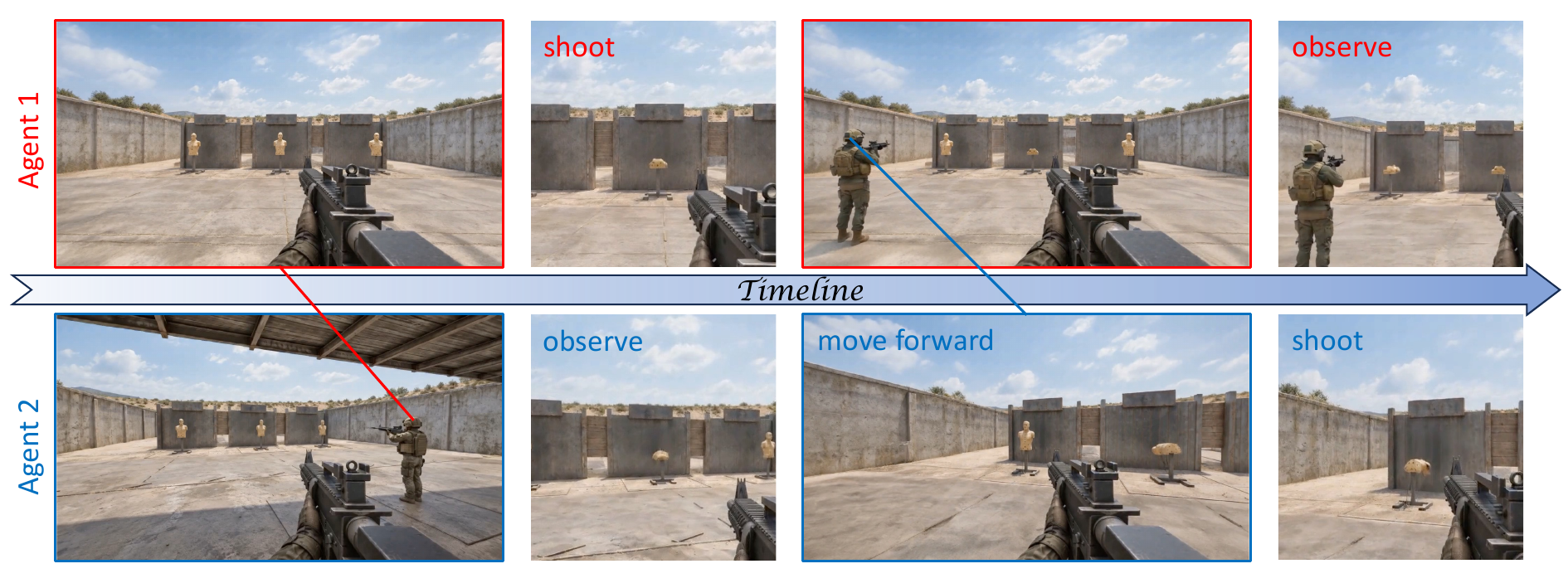}
    \vspace{-0.4cm}
    \captionsetup{font=small}
    \caption{We further explore multi-agent scenarios in our system. When Agent 1 takes down a target, the event is simultaneously observed by Agent 2. Likewise, when Agent 2 moves forward and shoots another target, the event is also observed by Agent 1.}
    \label{fig:comp3}
\end{figure}

\textbf{Multi-agent Scenarios.}
We additionally explore multi-Traveler (multi-agent) scenarios~\citep{wm_khora_2026,wm_multiplayer_rae_2026,wm_worldweaver_2026,wm_gamma_world_2026} and find that, with our proposed Observation as World Registration paradigm, the actions of one Traveler can modify the shared environment, while the resulting changes are simultaneously reflected in the observations of other agents. As shown in~\cref{fig:comp3}, when Agent 1 takes down a target, the event is observed by Agent 2. Likewise, when Agent 2 moves forward and takes down another target, the event is also observed by Agent 1. These results demonstrate the potential of our CoDeR as a promising framework for multi-agent world modeling.

\subsection{Quantitative Comparison}
We conduct a quantitative comparison with ABot-World, LingBot-World, and Matrix-Game 3.0 across 10 metrics on the WorldScore~\citep{wm_worldscore_2025} benchmark. The results are shown in~\cref{tab:quant}, where our method achieves state-of-the-art performance across all metrics compared with the other approaches.

\begin{table}[h]
    \centering
    \captionsetup{font=small}
    \caption{\textbf{Quantitative Comparison.} Our method outperforms all competing methods across all metrics.}
    \vspace{-0.3cm}
    \label{tab:quant}
    \resizebox{\linewidth}{!}{%
    \begin{tabular}{c|ccccccccccc}
        \toprule
        Methods & \makecell{Camera\\Ctrl} & \makecell{Object\\Ctrl} & \makecell{Content\\Align} & \makecell{3D\\Consist} & \makecell{Photo\\Consist} & \makecell{Style\\Consist} & \makecell{Subjective\\Qual} & \makecell{Motion\\Acc} & \makecell{Motion\\Mag} & \makecell{Motion\\Smooth} & \makecell{Average} \\
        \midrule
        ABot-World~\citep{wm_abot_world_0_2026} & 92.47 & 84.23 & 76.94 
        & 80.60 & 83.56 & 83.22
        & 39.95 & 50.33 & 26.89 & 74.21 
        & 69.24 \\
        Matrix-Game 3.0~\citep{wm_matrix_game_3_2026} & 96.72 & 84.61 & 83.12 
        & 81.10 & 86.96 & 83.42
        & 59.04 & 60.79 & 24.75 & 81.50
        & 74.20\\
        LingBot-World~\citep{wm_lingbot_world_2026} & 90.50 & 87.73 & 75.29 
        & 86.28 & 90.03 & 85.39
        & 65.71 & 59.48 & 27.93 & 79.60
        & 74.79\\
        \midrule
        \textbf{Ours} & \textbf{98.91} & \textbf{92.25} & \textbf{98.44} & \textbf{88.65} & \textbf{91.10} & \textbf{89.40} & 
        \textbf{69.08} & \textbf{82.76} & \textbf{84.92} & \textbf{83.31} & \textbf{87.88} \\
        \bottomrule
    \end{tabular}%
    }
\end{table}

\subsection{Ablation Study}
We conduct ablation studies on our Visual Cue Hacking strategy and Observation as World Registration paradigm using three metrics from WorldScore~\citep{wm_worldscore_2025} and four metrics from VBench~\citep{wm_vbench_2023} to evaluate their contributions to visual quality. As shown in~\cref{tab:ablate}, removing either Visual Cue Hacking or Observation as World Registration leads to degraded visual quality compared with the full model.

\begin{table}[h]
    \centering
    \captionsetup{font=small}
    \caption{\textbf{Ablation Study.} Both Visual Cue Hacking and Observation as World Registration contribute to improved visual quality across WorldScore and VBench metrics.}
    \label{tab:ablate}
    \vspace{-0.3cm}
    \resizebox{0.8\linewidth}{!}{%
    \begin{tabular}{c|ccccccc}
        \toprule
        \multirow{3}{*}{Methods}
        & \multicolumn{3}{c}{WorldScore}
        & \multicolumn{4}{c}{VBench} \\
        \cmidrule(lr){2-4} \cmidrule(lr){5-8}
        & \makecell{Content\\Align}
        & \makecell{Photo\\Consist}
        & \makecell{Subjective\\Qual}
        & \makecell{Imaging\\Quality}
        & \makecell{Aesthetic\\Quality}
        & \makecell{Subject\\Consistency}
        & \makecell{Dynamic\\Degree} \\
        \midrule
        w/o Vis. Cue Hack.
        & 82.48 & 89.21 & 49.05 & 63.52 & 66.18 & 87.54 & 97.29 \\
        w/o Obs. as WR.
        & 89.06 & 84.85 & 53.98 & 69.37 & 70.44 & 84.63 & 98.02 \\
        \midrule
        \textbf{Full Method}
        & \textbf{98.44} & \textbf{91.10} & \textbf{69.08} & \textbf{70.77} & \textbf{72.29} & \textbf{96.31} & \textbf{98.16} \\
        \bottomrule
    \end{tabular}%
    }
\end{table}




\section{Conclusion}
In this work, we introduced \textbf{CoDeR}, a new paradigm for world modeling that separates the underlying world from its visual realization. Instead of relying on video models to implicitly encode world dynamics, our framework constructs persistent and executable worlds through collaborative coding agents, while employing generative models to render perceptual observations. Extensive experiments demonstrate that this design not only broadens the capabilities of current world models, but also achieves state-of-the-art performance across diverse settings. More broadly, we hope this work encourages a shift from modeling worlds as sequences of observations toward constructing worlds as persistent computational systems that can be created, experienced, and continuously evolved.
And finally, every Traveler is, in essence, the God of Concepts.

\bibliography{iclr2027_conference}
\bibliographystyle{iclr2027_conference}

\newpage
\appendix
\section*{Appendix}
In this appendix, we first clarify how AI is used in this paper, followed by details of the user study, Visual Cue Hacking and evaluation. We then discuss the limitations of our method and outline directions for future work.

\subsection*{AI Use Statement}
We use GPT-6 Astra as the backbone model for both the Creator and Executors in our CoDeR, as described in~\cref{exp:whitebox}. The use of AI in these components constitutes a methodological design of our proposed framework rather than auxiliary assistance. In addition, we use AI-based tools to assist with language polishing and improve the clarity and readability of the manuscript. All technical content, experimental design, analysis, and conclusions are reviewed and finalized by the authors.

\subsection*{User Study}
\begin{figure}[h]
    \centering
   \includegraphics[width=0.8\linewidth]{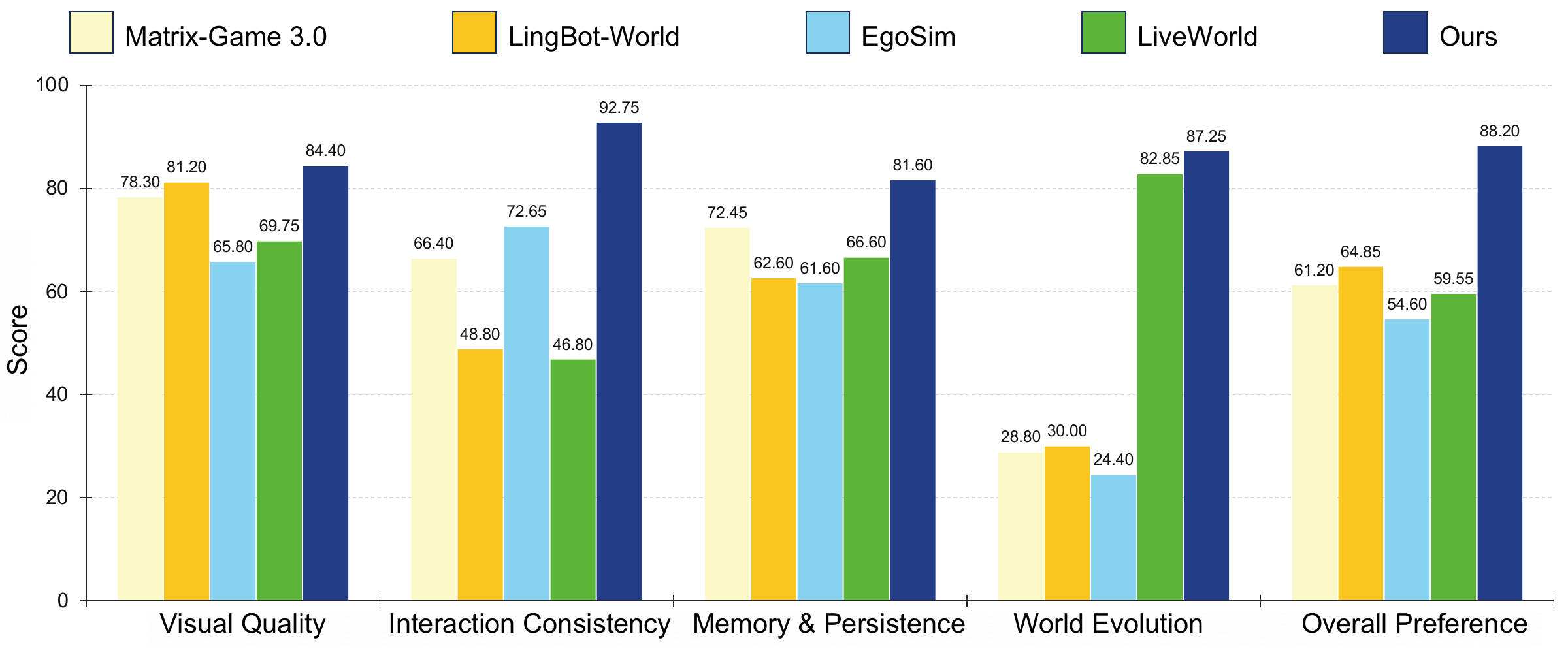}
    \captionsetup{font=small}
    \caption{\textbf{User Study.} Human evaluation across Visual Quality, Interaction Consistency, Memory and Persistence, World Evolution, and Overall Preference. Our method consistently achieves the highest scores across all evaluated dimensions.}
    \label{fig:user}
\end{figure}

To further evaluate our proposed CoDeR, we conduct a user study comparing it with Matrix-Game 3.0~\citep{wm_matrix_game_3_2026}, LingBot-World~\citep{wm_lingbot_world_2026}, EgoSim~\citep{wm_egosim_2026}, and LiveWorld~\citep{duan2026liveworld} across the following dimensions: Visual Quality, Interaction Consistency, Memory \& Persistence, World Evolution, and Overall Preference. Each participant is asked to score the videos generated by different methods along each dimension. As shown in~\cref{fig:user}, we collect 30 valid questionnaires, and the results show that our method consistently outperforms the competing approaches across all evaluated dimensions.

\subsection*{Visual Cue Hacking}
\begin{figure}[t]
    \centering
   \includegraphics[width=0.9\linewidth]{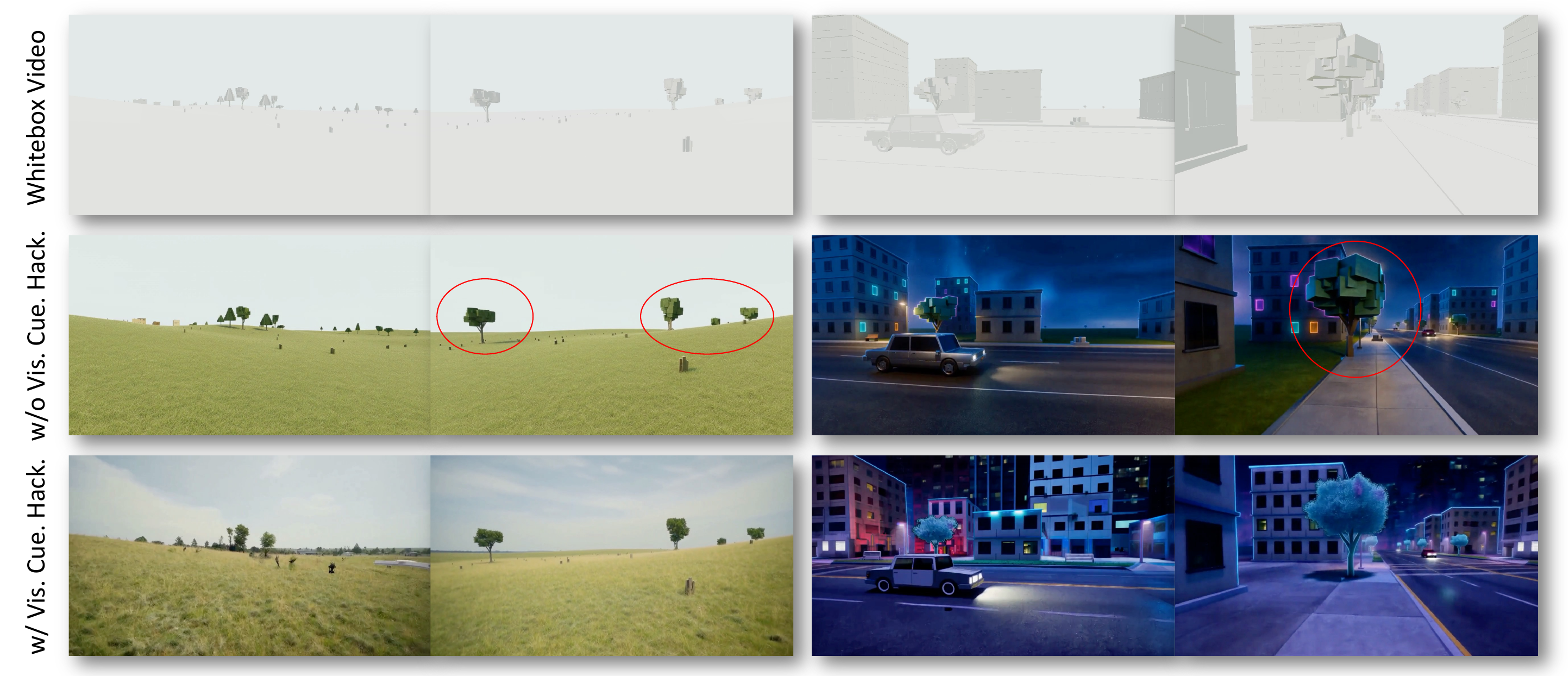}
    \captionsetup{font=small}
    \caption{\textbf{Effect of Visual Cue Hacking.} Directly conditioning on whitebox renderings produces monotonous, render-like videos with rigid edges, whereas our Visual Cue Hacking strategy encourages the model to use the whitebox input as a structural cue, resulting in richer appearance and higher visual quality.}
    \label{fig:vch}
\end{figure}

We further illustrate the effect of our Visual Cue Hacking strategy in~\cref{fig:vch}. We compare videos generated by directly conditioning on the whitebox renderings with those produced using Visual Cue Hacking. As shown in~\cref{fig:vch}, directly rendering from whitebox videos tends to produce visually monotonous, render-like results with overly sharp and rigid edges, as the generation model closely follows the low-level geometry of the white-box inputs. In contrast, Visual Cue Hacking effectively adapts the model to the whitebox domain while preventing it from overfitting to these artificial boundaries. As a result, the model treats the whitebox video primarily as a structural cue rather than a pixel-level rendering target, enabling substantially higher visual quality with richer appearance and more natural details.

\subsection*{Evaluation Details}
\begin{figure}[h]
    \centering
   \includegraphics[width=0.9\linewidth]{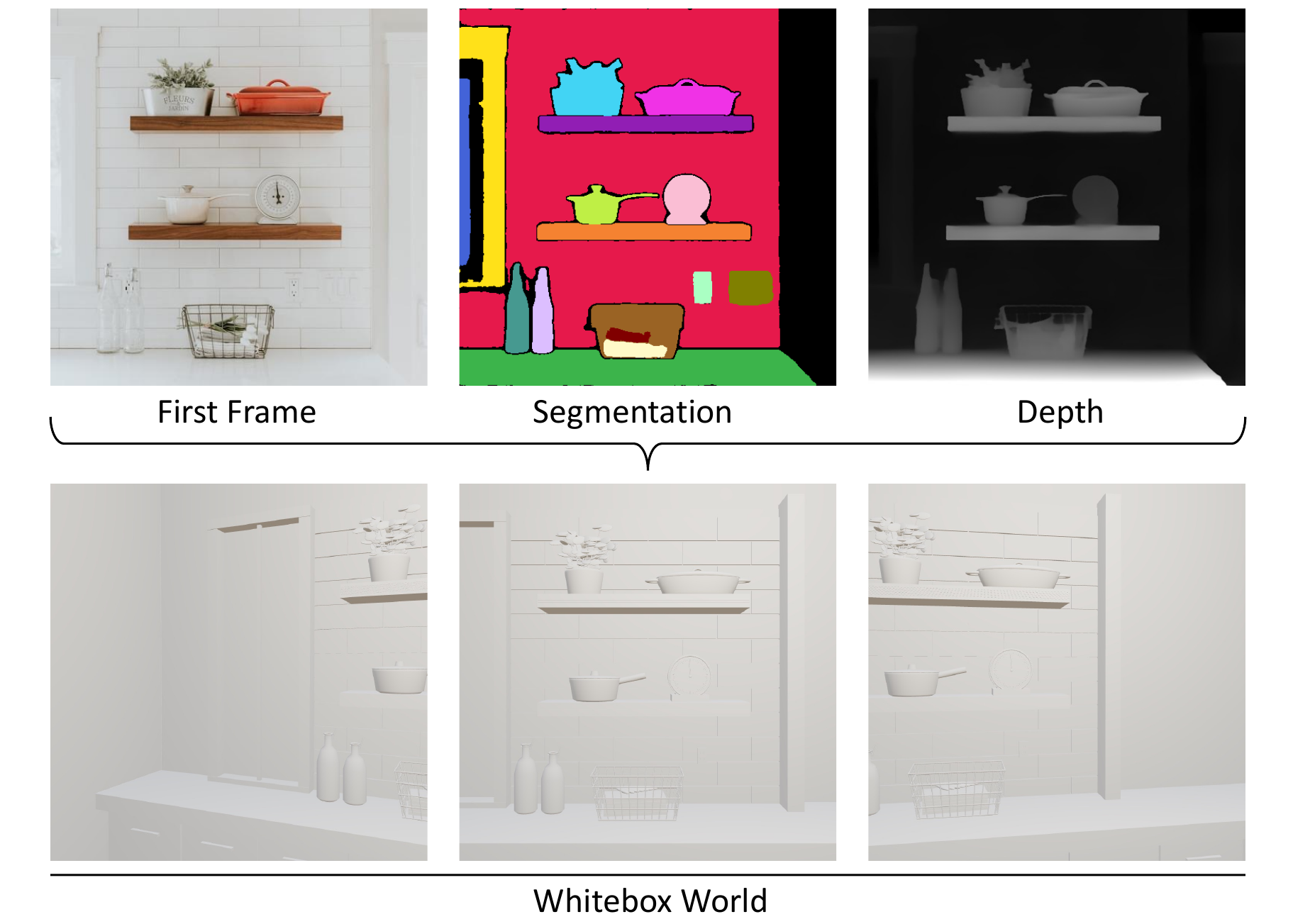}
    \captionsetup{font=small}
    \caption{Given the first frame, we obtain its segmentation map and depth map, and then reconstruct a corresponding whitebox world.}
    \label{fig:evalDet}
\end{figure}

Since our method cannot be directly adapted to the input formats of many existing benchmarks, we first construct a corresponding whitebox world from the benchmark inputs and then sample videos within this reconstructed environment for evaluation. Specifically, given a first frame, we first use SAM 3~\citep{carion2026sam} to segment the scene into its constituent objects and employ Depth Anything V2~\citep{yang2024depth} to estimate the corresponding depth map. We then feed the first frame, segmentation map, and depth map into GPT-6 Astra to construct the whitebox world, as illustrated in~\cref{fig:evalDet}.

Note that the reconstructed whitebox world is not required to be perfectly aligned with the first frame. In practice, discrepancies may arise in object geometry, shape, or fine-grained scene structure. Nevertheless, we find that conditioning the generation model on the original first frame effectively anchors the visual content to the reconstructed world. Even when the geometry of an object is only approximately matched, the generated appearance can still be aligned with its corresponding location in the whitebox world, allowing the reconstructed environment to serve as a reliable structural scaffold for benchmark evaluation.

\subsection*{Limitations}
\begin{figure}[h]
    \centering
   \includegraphics[width=0.9\linewidth]{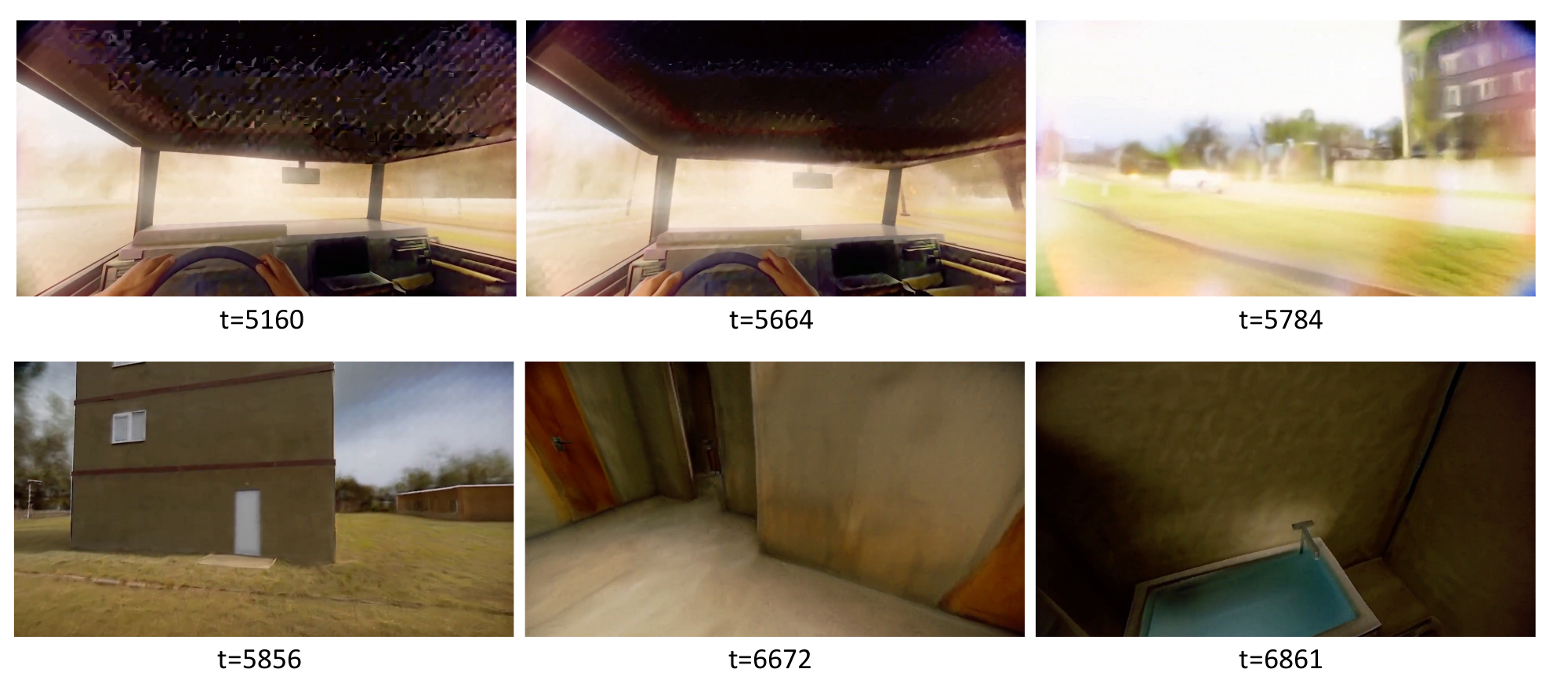}
    \captionsetup{font=small}
    \caption{\textbf{Limitation of Long-Horizon Video Generation.} Our Artist model is trained with a fixed 124-frame window and generates long videos through chunk-based rollout. Noticeable visual degradation emerges due to accumulated generation errors.}
    \label{fig:long}
\end{figure}

\textbf{Extreme-Long Video Generation.}
Our current Artist model is fine-tuned with a fixed temporal window of 124 frames and does not undergo additional post-training specifically designed for long-horizon video generation. To generate longer sequences, we adopt a chunk-based generation strategy and concatenate multiple clips autoregressively. As a result, visual errors and distribution shifts can gradually accumulate over time, inevitably leading to quality degradation in extremely long rollouts. As shown in~\cref{fig:long}, at approximately 5,000 frames, noticeable degradation appears in the generated video. We believe that incorporating dedicated long-video post-training or more effective temporal memory mechanisms could further improve long-horizon generation quality.

\begin{figure}[h]
    \centering
   \includegraphics[width=0.9\linewidth]{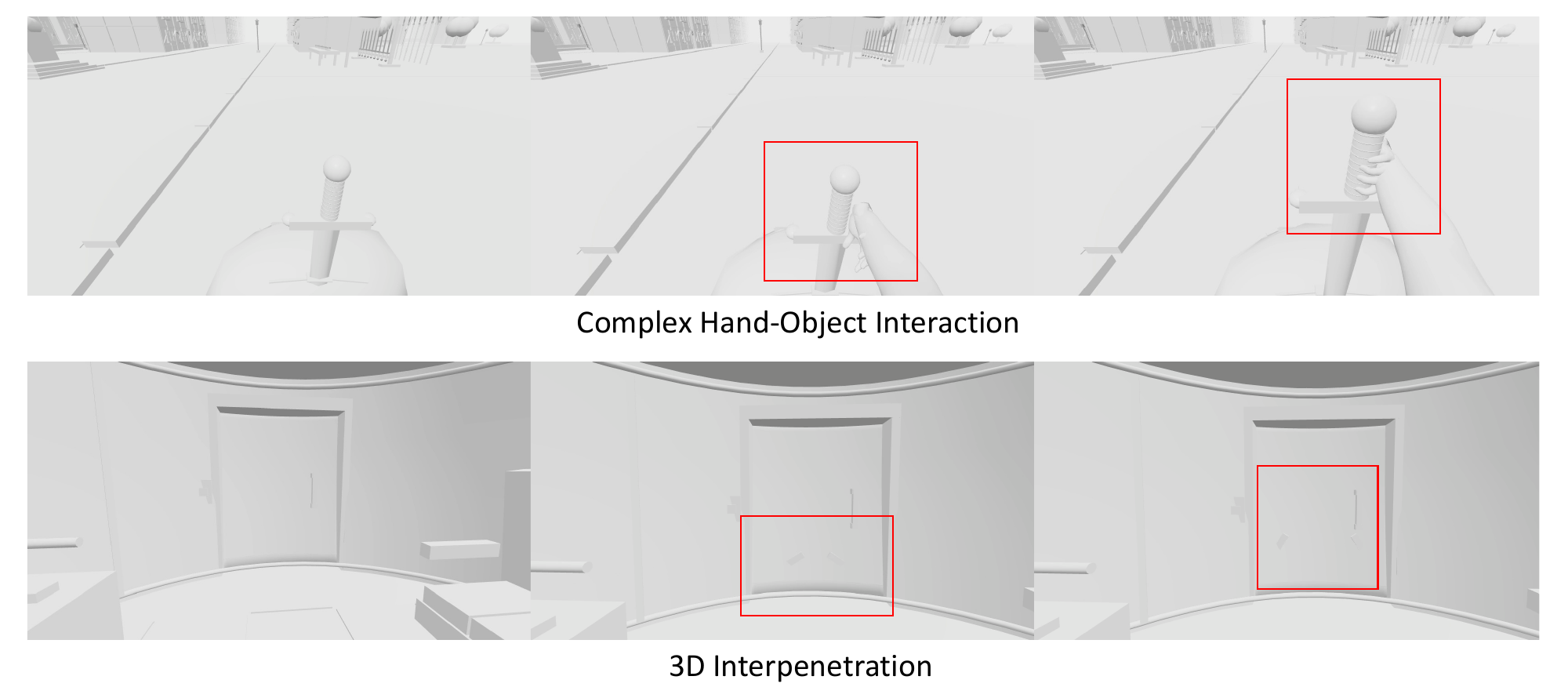}
    \captionsetup{font=small}
    \caption{\textbf{Limitations of Whitebox Modeling.} Our current whitebox world may exhibit failure cases in complex hand-object interactions and 3D geometry, including severely distorted hand poses and object interpenetration.}
    \label{fig:whiteM}
\end{figure}

\textbf{Whitebox Modeling.}
Our framework also inherits limitations from the construction and simulation of the whitebox world. In particular, accurately modeling complex hand-object interactions remains challenging, where the reconstructed or simulated hand poses may exhibit severe geometric distortion. In addition, imperfect scene geometry or collision handling can occasionally result in 3D interpenetration between objects. Representative failure cases are shown in~\cref{fig:whiteM}. These limitations suggest that more accurate geometry reconstruction, articulated object modeling, and physically grounded interaction simulation could further improve the fidelity of the underlying world representation.

\subsection*{Future work}
\begin{figure}[h]
    \centering
   \includegraphics[width=0.9\linewidth]{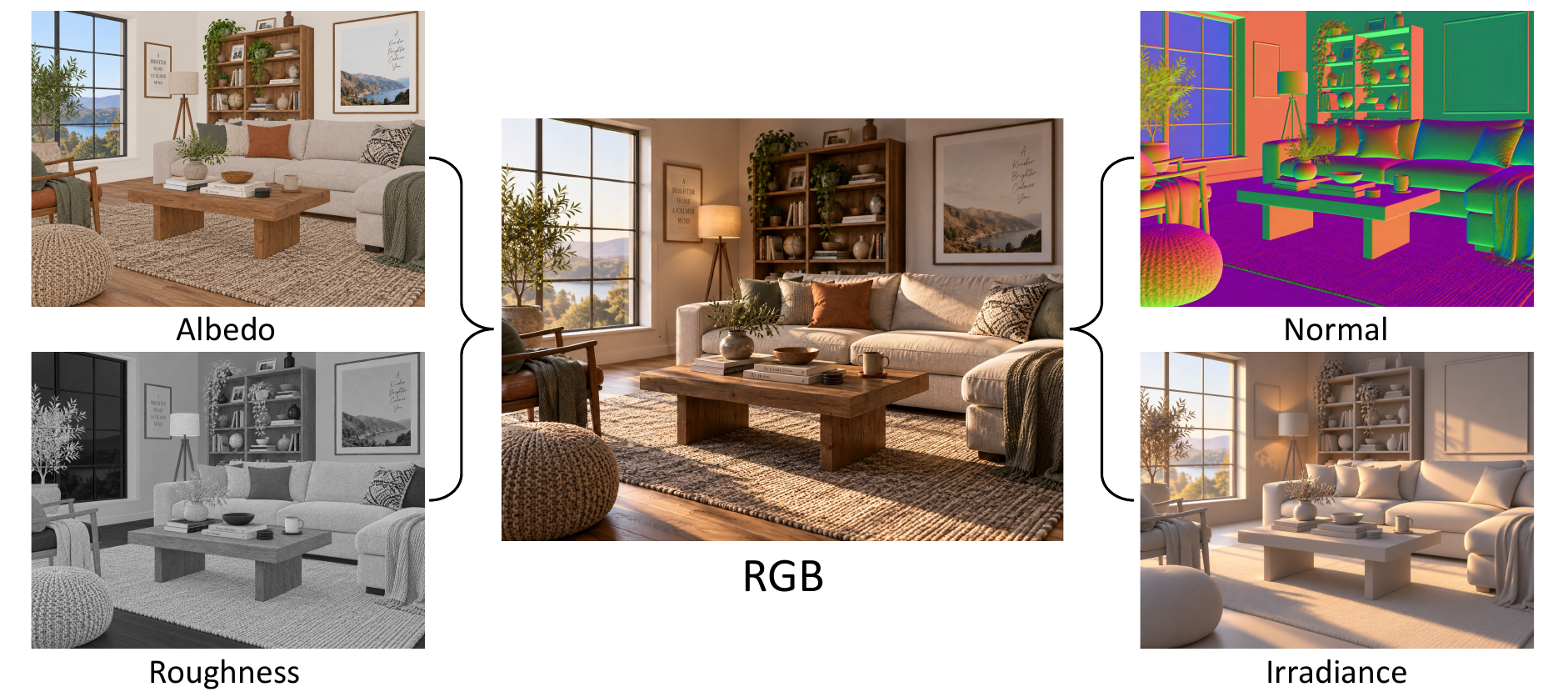}
    \captionsetup{font=small}
    \caption{\textbf{Towards Disentangled Visual Generation.} Future Artist models could decompose RGB observations into structured visual representations, including Albedo, Normal, Roughness, and Irradiance, enabling more explicit scene registration, editing, and re-rendering within the CoDeR.}
    \label{fig:rgbx}
\end{figure}

Beyond improving long-horizon video generation as discussed above, an important direction for future work is to further disentangle the visual generation process of the Artist model. Our current Artist directly generates complete RGB observations, in which geometry, material properties, illumination, and appearance are implicitly entangled. To better align with the paradigm proposed in this work, future Artist models could instead generate decomposed visual representations, such as Albedo, Normal, Roughness, and Irradiance maps, following recent progress in material- and lighting-aware image decomposition and synthesis~\citep{zeng2024rgb}. As illustrated in~\cref{fig:rgbx}, such factorized representations could provide more explicit and reusable scene information, making generated observations easier to register, update, and re-render within the underlying world. We believe this direction could further strengthen our Observation as World Registration paradigm by turning observations from monolithic RGB frames into structured visual states that can be more faithfully integrated back into the world.

\end{document}